\documentclass[conference, flushend]{iaria}
\IEEEoverridecommandlockouts
\usepackage{hyperref}
\usepackage{amsmath,amssymb,amsfonts}
\usepackage{algorithmic}
\usepackage{graphicx}
\usepackage{textcomp}
\usepackage{xcolor}
\usepackage{times}
\usepackage[nolist]{acronym}
\usepackage{color}
\usepackage{booktabs}
\usepackage{tabularx}
\usepackage{multirow}
\usepackage[normalem]{ulem}
\usepackage{biblatex}
\usepackage{ccicons}
\usepackage[shortcuts]{extdash}

\acrodef{LLM}[LLM]{Large Language Model}
\acrodef{GPT}[GPT]{Generative Pre-trained Transformer}
\acrodef{RAG}[RAG]{Retrieval Augmented Generation}
\acrodef{ARC}[ARC]{AI2 Reasoning Challenge}
\acrodef{MMLU}[MMLU]{Massive Multitask Language Understanding}
\acrodef{MTEB}[MTEB]{Massive Text Embedding Benchmark} 
\acrodef{MoE}[MoE]{Mixture-of-Experts}
\acrodef{AI}[AI]{Artificial Intelligence}

\makeatletter
\def\ps@IEEEtitlepagestyle{
  \def\@oddfoot{\mycopyrightnotice}
  \def\@evenfoot{}
}
\def\mycopyrightnotice{
  {\footnotesize
    \begin{minipage}{0.8\textwidth}
    \centering
    Please cite as: \fullcite{selfref}.
    \end{minipage}
  }
}
\makeatother

\DeclareBibliographyCategory{selfref}
\addtocategory{selfref}{selfref}

\title{\bfseries\Large Comparison of Large Language Models for Deployment Requirements}

\author{\IEEEauthorblockN{Alper Yaman\IEEEauthorrefmark{2}\IEEEauthorrefmark{1},
Jannik Schwab\IEEEauthorrefmark{2},
Christof Nitsche\IEEEauthorrefmark{2},
Abhirup Sinha\IEEEauthorrefmark{2} and
Marco Huber\IEEEauthorrefmark{2}}\IEEEauthorblockA{\IEEEauthorrefmark{2}Department Cyber Cognitive Intelligence\\Fraunhofer Institute for Manufacturing Engineering and Automation IPA, Stuttgart, Germany\\Email: firstname.lastname@ipa.fraunhofer.de}
}

\begin{document}
\maketitle

\begin{abstract}
\acp{LLM}, such as \acp{GPT} are revolutionizing the generation of human-like text, producing contextually relevant and syntactically correct content. Despite challenges like biases and hallucinations, these \ac{AI} models excel in tasks, such as content creation, translation, and code generation. Fine-tuning and novel architectures, such as Mixture of Experts (MoE), address these issues. Over the past two years, numerous open-source foundational and fine-tuned models have been introduced, complicating the selection of the optimal LLM for researchers and companies regarding licensing and hardware requirements. To navigate the rapidly evolving LLM landscape and facilitate LLM selection, we present a comparative list of foundational and domain-specific models, focusing on features, such as release year, licensing, and hardware requirements. This list is published on GitLab and will be continuously updated.
\end{abstract}


\section{Introduction}
\acfp{LLM} like \acf{GPT} are advanced \acf{AI} models designed to generate human-like text in response to the input they receive. These foundational models differ in underlying architecture, training procedures, and training data. They are trained on vast datasets containing a diverse range of internet text. They work by predicting the next word in a sequence, making them proficient at generating coherent sentences, and even writing poems or computer scripts. 

The ability of \acp{LLM} to generate contextually relevant and syntactically correct text has revolutionized fields, such as content creation, customer service, and software development. \acp{LLM} are also integral in developing tools for language translation, summarization, and question-answering systems, enhancing accessibility and efficiency. Furthermore, they contribute significantly to research in natural language understanding and generation, pushing the boundaries of AI's capabilities in understanding complex language constructs. 

However, \acp{LLM} can produce hallucinations, i.e., generating biased or incorrect information, which raises major concerns about their use in sensitive areas like law and healthcare. To address these drawbacks, pre-trained models are fine-tuned with domain-specific, task-specific corpora or instructions. Another method is \ac{MoE} \acp{LLM}, where a set of \acp{LLM} (experts) attend to different parts of the input space. This concept is similar to ensemble methods in traditional machine learning, where the outputs from a set of models are voted to provide a single, more accurate outcome. 

Despite these challenges, \acp{LLM} continue to be a pivotal area of research and development, resulting in a vast number of scientific articles. New jargon has rapidly emerged concerning the operation and evaluation of \acp{LLM}, including terms, such as prompt engineering, instruction-based fine-tuning \cite{zhang2024instruction}, and \ac{RAG} \cite{gao2024retrievalaugmented}. Additionally, the evaluation of the accuracy and performance of \acp{LLM} has been questioned, leading to the proposal of various metrics \cite{chang.2024}. Multiple surveys have been published that provide comprehensive insights into recent advancements \cite{zhao2023survey}\cite{huang2023reasoning}, discuss evaluation metrics from the perspective of explainability \cite{zhao2024_exp}, and aim to align \acp{LLM} with human expectations \cite{wang2023aligning}. 

In addition to closed-source cloud-based \acp{LLM} like ChatGPT, numerous models have been uploaded to HuggingFace for community use. However, these models vary in features, such as model size, embedding dimensions, and max token count, with details listed on platforms like HuggingFace and Github, and surveys \cite{zhao2023survey}\cite{huang2023reasoning}. This variability makes it challenging for companies and researchers to select an \ac{LLM} that meets specific requirements, particularly when the model is intended for local deployment. 

The aim of this study is to provide a comparative list of foundational and domain-specific models to support companies and researchers in selecting \acp{LLM}. In section \ref{sec: relatedwork}, we explain some of the existing \acp{LLM} lists, their content, and the parameters with which they are compared. In section \ref{sec: proposedwork}, we detail which models are selected and which features are compared. In section \ref{sec: results}, basic statistics about the listed \acp{LLM} are provided, and a part of the comparison list is shown. In section \ref{sec: conclusion}, further information is given about how the list will be maintained in the feature and the limitations of this study. 

\section{Related Work}
\label{sec: relatedwork}
%\todo{Abhirup: N: number of models here} 
As of May 2024 when this study was performed, HuggingFace had approximately 65 pre-trained \acp{LLM} for text generation tasks pertaining to the English language. Additionally, many fine-tuned models, based on the pre-trained models, have been uploaded to HuggingFace \cite{open-llm-leaderboard}. This platform has a couple of leaderboards that compare the fine-tuned models using a framework for few-shot language model evaluation \cite{eval-harness}. The Open LLM Leaderboard compares models regarding their type, architecture, model precision, average accuracy, as well as accuracy values calculated separately using various datasets and benchmarks. Another leaderboard is \ac{MTEB} Leaderboard illustrating the model size, memory usage, embedding dimensions, max tokens, average overall accuracy from 56 datasets, and average accuracies for classification, clustering, pair classification, reranking, retrieval, STS, and summarization from 12, 11, 3, 4, 15, 10, and 1 datasets, respectively \cite{muennighoff2022mteb}. A total of 281 models are compared with 159 datasets for 113 languages. LMSYS Chatbot Arena Leaderboard is a crowdsourced open platform to evaluate \acp{LLM} \cite{chiang2024chatbot}. As of April 24, 2024, 91 models were evaluated using 800,000 human pairwise comparisons to rank them with the Bradley-Terry model \cite{bradley-terry}. Additionally, there are some Github repositories \cite{eugeneyan-github} and websites \cite{sapling-web} that provide rough comparisons. Note that none of these leaderboards provides comprehensive details when companies and researchers encounter technical challenges when they deploy an \ac{LLM} on their own hardware.

These tables compare the success scores of the \acp{LLM} along with their basic information (e.g., type and architecture) but omit the requirements for deployment. Including these requirements is essential to streamline the feasibility analysis process when selecting the most suitable \ac{LLM}. Our comparison list addresses these needs by providing information on hardware and licensing requirements.

\section{Proposed Work}
\label{sec: proposedwork}
%(\todo{Abhirup: some words about selection criteria}). 
In this study, we created an extensive comparison list of \acp{LLM} for researchers and companies to simplify \ac{LLM} selection. Since there are numereous fine-tuned models, we primarily focused on covering base foundational \acp{LLM}, as much as possible. Nevertheless, some existing domain-specific (e.g., in the medical domain) fine-tuned models were included. We then defined the model features that help users to select the correct \ac{LLM}. To easily distinguish between different \acp{LLM}, we provided both \ac{LLM} names and families together with the model features, such as release year, license types, and hardware requirements.

The outcome of this study, in the form of a comparison table, is published on a GitLab page for community use. Since new \acp{LLM} and their derivatives are continually being developed, this is an ongoing effort, and the GitLab page will be updated regularly\protect\cite{GenAIModelOverview}.
%\todo{figure, flow chart. Any suggestion?}

\subsection{Model Selection}
%\todo{@Abhirup: Why did we select the LLMs on our list? Statistics about the LLMs, like how many foundation models exist, LLM families, and their differences.}
We selected 108 \acp{LLM} based on the criteria of being open-source and having been published in or after 2023. Approximately 20 of them are foundational \acp{LLM}, such as Mistral, LLaMA-2, LLaMA-3, Code LLaMA, Gemma, RecurrentGemma, Falcon, Dolly, etc. Some fine-tuned \acp{LLM}, such as BioMistral, Meditron, and Medicine-LLM, as well as several \ac{MoE} \acp{LLM} (e.g., Mixtral, Grok-1, and DBRX) were included.

\subsection{Model Features}
%\todo{@Abhirup: Explanation of the features (columns in the table), Basic info about the licences (what they mean) }
We included information on \ac{LLM} families and the versions existing within the \ac{LLM} families. The sizes (i.e., number of parameters) and release dates were listed to track the gradual development in this field.

Furthermore, we also investigated the commercial aspects of the listed open-source \acp{LLM} and listed the license information. Since understanding the licenses can be difficult for readers, in another column, we clarified if the licenses allow for commercial usage of the model (with or without any restrictions) or not. 

In addition, we included information on minimum memory requirements (RAM and vRAM) and required disk space for complete fine-tuning and inference. Note that these requirements are applicable for loading the 5-bit quantized versions of the models. Loading models with full-precision floating point numbers usually requires twice or four times more memory relative to their parameters.

\section{Results}
\label{sec: results}
%\todo{Resulting table is here}
%CPU Inference Speed (Q: What are the medical reasons behind Malaria?)- 
%BioMistral-7B : No Response in CPU inference;
%Meditron-7B : around 2 tokens per second;
%Gemma-2B : around 6 tokens per second;
%LLaMA-3-8B : around 2 tokens per second;
%TinyLLaMA : around 12 tokens per second;

A small subset of our resulting table is shown in Table \ref{tab:llm-table} \cite{GenAIModelOverview}. The information on \acp{LLM}, along with their families, license, and memory requirements is listed to provide a quick overview of the \acp{LLM} for the specific needs and use cases of researchers and companies.

\begin{table*}[tbp]
\centering
\caption{A Snapshot of the Table of Current Open-Source \acp{LLM}}
\label{tab:llm-table}
\resizebox{\textwidth}{!}{%
\begin{tabular}{@{}|c|c|c|c|c|c|c|ccc|@{}}
\toprule
\textbf{} &
  \textbf{} &
  \textbf{} &
  \textbf{} &
  \textbf{} &
  \textbf{} &
  \textbf{Fine-tuning} &
  \multicolumn{3}{c|}{\textbf{Inference}} \\ \midrule
\textbf{Family} &
  \textbf{Name} &
  \textbf{Release Year} &
  \textbf{Size (B Parameters)} &
  \textbf{License type} &
  \textbf{Commercial Usage} &
  \textbf{Min. GB GPU} &
  \multicolumn{1}{c|}{\textbf{Min. GB RAM}} &
  \multicolumn{1}{c|}{\textbf{Min. GB GPU}} &
  \textbf{Min. GB Disk Space} \\ \midrule
\multirow{2}{*}{Code} &
  Code-13B &
  Dec 23 &
  13 &
  CC-BY-NC-ND 4.0 &
  No &
  26 &
  \multicolumn{1}{c|}{11.73} &
  \multicolumn{1}{c|}{5.4} &
  9.23 \\ \cmidrule(l){2-10} 
 &
  Code-33B &
  Dec 23 &
  33 &
  CC-BY-NC-ND 4.0 &
  No &
  66 &
  \multicolumn{1}{c|}{25.55} &
  \multicolumn{1}{c|}{13.5} &
  23.05 \\ \midrule
\multirow{9}{*}{CodeLLaMA} &
  7B &
  Aug 23 &
  7 &
  LLaMA-2 &
  Partial &
  14 &
  \multicolumn{1}{c|}{7.28} &
  \multicolumn{1}{c|}{2.8} &
  4.78 \\ \cmidrule(l){2-10} 
 &
  7B-Instruct &
  Aug 23 &
  7 &
  LLaMA-2 &
  Partial &
  14 &
  \multicolumn{1}{c|}{7.28} &
  \multicolumn{1}{c|}{2.8} &
  4.78 \\ \cmidrule(l){2-10} 
 &
  7B-Python &
  Aug 23 &
  7 &
  LLaMA-2 &
  Partial &
  14 &
  \multicolumn{1}{c|}{7.28} &
  \multicolumn{1}{c|}{2.8} &
  4.78 \\ \cmidrule(l){2-10} 
 &
  13B &
  Aug 23 &
  13 &
  LLaMA-2 &
  Partial &
  26 &
  \multicolumn{1}{c|}{11.73} &
  \multicolumn{1}{c|}{5.4} &
  9.23 \\ \cmidrule(l){2-10} 
 &
  13B-Instruct &
  Aug 23 &
  13 &
  LLaMA-2 &
  Partial &
  26 &
  \multicolumn{1}{c|}{11.73} &
  \multicolumn{1}{c|}{5.4} &
  9.23 \\ \cmidrule(l){2-10} 
 &
  13B-Python &
  Aug 23 &
  13 &
  LLaMA-2 &
  Partial &
  26 &
  \multicolumn{1}{c|}{11.73} &
  \multicolumn{1}{c|}{5.4} &
  9.23 \\ \cmidrule(l){2-10} 
 &
  34B &
  Aug 23 &
  34 &
  LLaMA-2 &
  Partial &
  68 &
  \multicolumn{1}{c|}{26.84} &
  \multicolumn{1}{c|}{14.2} &
  23.84 \\ \cmidrule(l){2-10} 
 &
  34B-Instruct &
  Aug 23 &
  34 &
  LLaMA-2 &
  Partial &
  68 &
  \multicolumn{1}{c|}{26.84} &
  \multicolumn{1}{c|}{14.2} &
  23.84 \\ \cmidrule(l){2-10} 
 &
  34B-Python &
  Aug 23 &
  34 &
  LLaMA-2 &
  Partial &
  68 &
  \multicolumn{1}{c|}{26.84} &
  \multicolumn{1}{c|}{14.2} &
  23.84 \\ \midrule
\multirow{8}{*}{LLaMA-2} &
  7B &
  Jul 23 &
  7 &
  LLaMA-2 &
  Partial &
  14 &
  \multicolumn{1}{c|}{7.28} &
  \multicolumn{1}{c|}{2.8} &
  4.78 \\ \cmidrule(l){2-10} 
 &
  7B-Chat &
  Jul 23 &
  7 &
  LLaMA-2 &
  Partial &
  14 &
  \multicolumn{1}{c|}{7.28} &
  \multicolumn{1}{c|}{2.8} &
  4.78 \\ \cmidrule(l){2-10} 
 &
  7B-Coder &
  Dec 23 &
  7 &
  LLaMA-2 &
  Partial &
  14 &
  \multicolumn{1}{c|}{7.28} &
  \multicolumn{1}{c|}{2.8} &
  4.78 \\ \cmidrule(l){2-10} 
 &
  13B &
  Jul 23 &
  13 &
  LLaMA-2 &
  Partial &
  26 &
  \multicolumn{1}{c|}{11.73} &
  \multicolumn{1}{c|}{5.4} &
  9.23 \\ \cmidrule(l){2-10} 
 &
  13B-Chat &
  Jul 23 &
  13 &
  LLaMA-2 &
  Partial &
  26 &
  \multicolumn{1}{c|}{11.73} &
  \multicolumn{1}{c|}{5.4} &
  9.23 \\ \cmidrule(l){2-10} 
 &
  70B &
  Jul 23 &
  13 &
  LLaMA-2 &
  Partial &
  140 &
  \multicolumn{1}{c|}{51.25} &
  \multicolumn{1}{c|}{29.3} &
  48.75 \\ \cmidrule(l){2-10} 
 &
  70B-Chat &
  Jul 23 &
  70 &
  LLaMA-2 &
  Partial &
  140 &
  \multicolumn{1}{c|}{51.25} &
  \multicolumn{1}{c|}{29.3} &
  48.75 \\ \midrule 
Med42 &
  70B &
  Nov 23 &
  70 &
  Med42 &
  No &
  140 &
  \multicolumn{1}{c|}{51.25} &
  \multicolumn{1}{c|}{29.3} &
  48.75 \\ \midrule
\multirow{2}{*}{Starling LM} &
  7B-Alpha &
  Nov 23 &
  7 &
  CC-BY-NC 4.0 &
  No &
  14 &
  \multicolumn{1}{c|}{7.63} &
  \multicolumn{1}{c|}{2.7} &
  5.13 \\ \cmidrule(l){2-10} 
 &
  Alpha 8X7B MoE &
  Dec 23 &
  47 &
  CC-BY-NC 4.0 &
  No &
  94 &
  \multicolumn{1}{c|}{34.73} &
  \multicolumn{1}{c|}{17.3} &
  32.23 \\ \midrule
\multirow{4}{*}{WizardLM} &
  7B-v1.0 &
  Apr 23 &
  7 &
  Non-commercial &
  No &
  14 &
  \multicolumn{1}{c|}{7.28} &
  \multicolumn{1}{c|}{2.8} &
  4.78 \\ \cmidrule(l){2-10} 
 &
  13B-v1.2 &
  Jul 23 &
  13 &
  LLaMA-2 &
  Partial &
  26 &
  \multicolumn{1}{c|}{11.73} &
  \multicolumn{1}{c|}{5.4} &
  9.23 \\ \cmidrule(l){2-10} 
 &
  30B-v1.0 &
  Jun 23 &
  30 &
  Non-commercial &
  No &
  60 &
  \multicolumn{1}{c|}{25.55} &
  \multicolumn{1}{c|}{13.5} &
  23.05 \\ \cmidrule(l){2-10} 
 &
  70B-v1.0 &
  Aug 23 &
  70 &
  Non-commercial &
  No &
  140 &
  \multicolumn{1}{c|}{51.25} &
  \multicolumn{1}{c|}{29.3} &
  48.75 \\ \midrule
\multirow{3}{*}{Zephyr} &
  3B &
  Nov 23 &
  3 &
  StabilityAI Non-Commercial Research Community License &
  No &
  6 &
  \multicolumn{1}{c|}{4.49} &
  \multicolumn{1}{c|}{1.2} &
  1.99 \\ \cmidrule(l){2-10} 
 &
  7B-Alpha &
  Oct 23 &
  7 &
  MIT &
  Yes &
  14 &
  \multicolumn{1}{c|}{7.63} &
  \multicolumn{1}{c|}{2.7} &
  5.13 \\ \cmidrule(l){2-10} 
 &
  7B-Beta &
  Oct 23 &
  7 &
  MIT &
  Yes &
  14 &
  \multicolumn{1}{c|}{7.63} &
  \multicolumn{1}{c|}{2.7} &
  5.13 \\ \midrule
\multirow{4}{*}{BioMistral} &
  7B &
  Feb 24 &
  7 &
  Apache 2.0 &
  Yes &
  14 &
  \multicolumn{1}{c|}{7.63} &
  \multicolumn{1}{c|}{2.7} &
  5.13 \\ \cmidrule(l){2-10} 
 &
  7B-DARE &
  Feb 24 &
  7 &
  Apache 2.0 &
  Yes &
  14 &
  \multicolumn{1}{c|}{7.63} &
  \multicolumn{1}{c|}{2.7} &
  5.13 \\ \cmidrule(l){2-10} 
 &
  7B-TIES &
  Feb 24 &
  7 &
  Apache 2.0 &
  Yes &
  14 &
  \multicolumn{1}{c|}{7.63} &
  \multicolumn{1}{c|}{2.7} &
  5.13 \\ \cmidrule(l){2-10} 
 &
  7B-SLERP &
  Feb 24 &
  7 &
  Apache 2.0 &
  Yes &
  14 &
  \multicolumn{1}{c|}{7.63} &
  \multicolumn{1}{c|}{2.7} &
  5.13 \\ \midrule
TinyLLaMA &
  1.1B-Chat-v1.0 &
  Jan 2024 &
  1.1 &
  Apache 2.0 &
  Yes &
  2.2 &
  \multicolumn{1}{c|}{3.28} &
  \multicolumn{1}{c|}{0.5} &
  0.78 \\ \bottomrule
\end{tabular}%
}
%\raggedright
%Full Table at: \url{https://technology-project-aimv-projects-generative-ai-54af1e2b8cbbab0a.pages.fraunhofer.de/}
\end{table*}

\begin{figure}[htbp]
  \centering
  \resizebox{\columnwidth}{!}{\includegraphics{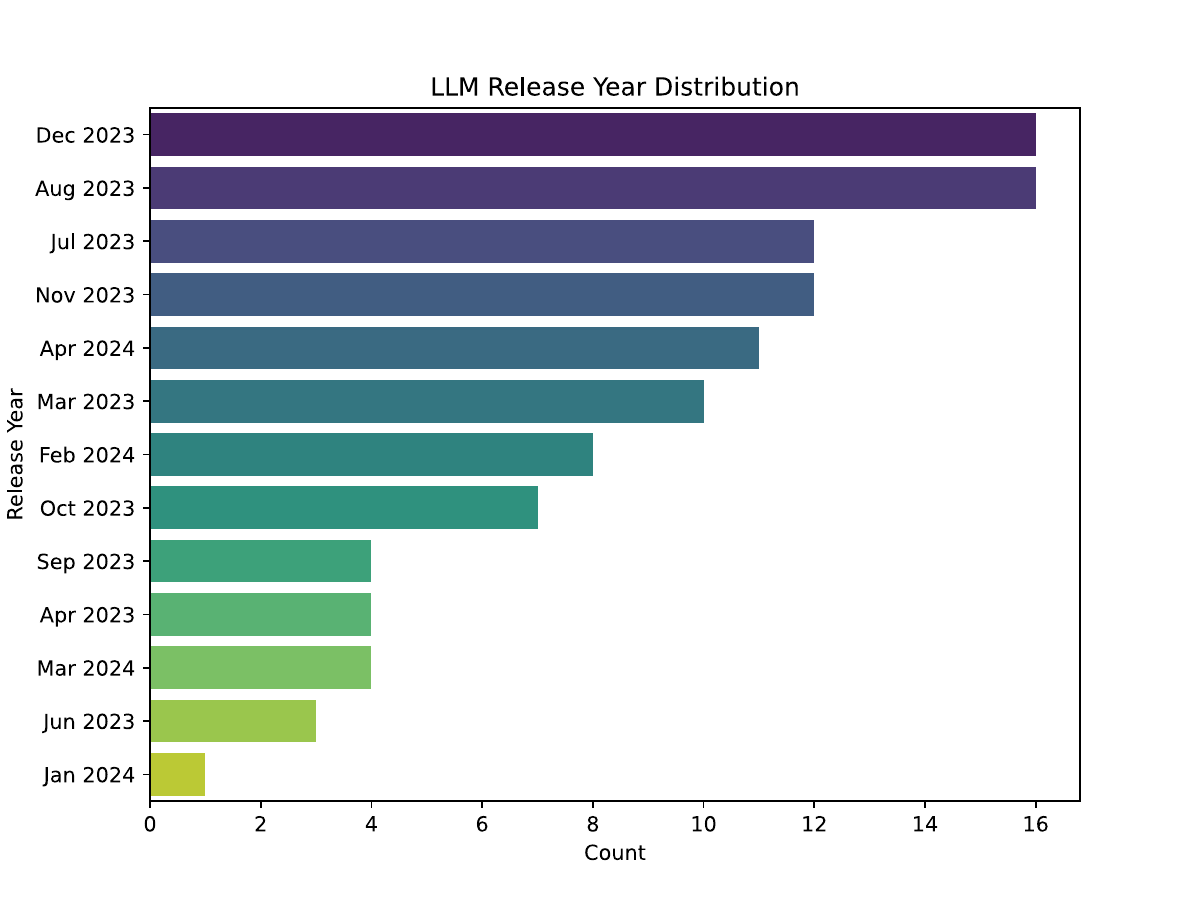}}
  \vspace{-20pt}
  \caption{Release Year Distribution of Listed \acp{LLM}}
  \label{fig:release_year_distribution}
\end{figure}

Figure \ref{fig:release_year_distribution} shows the distribution of release date, indicating that; most of the \acp{LLM} we listed were released in 2023. Note that the most recent \acp{LLM} on our list were released in April 2024.  

\begin{figure}[htbp]
  \centering
  \resizebox{\columnwidth}{!}{\includegraphics{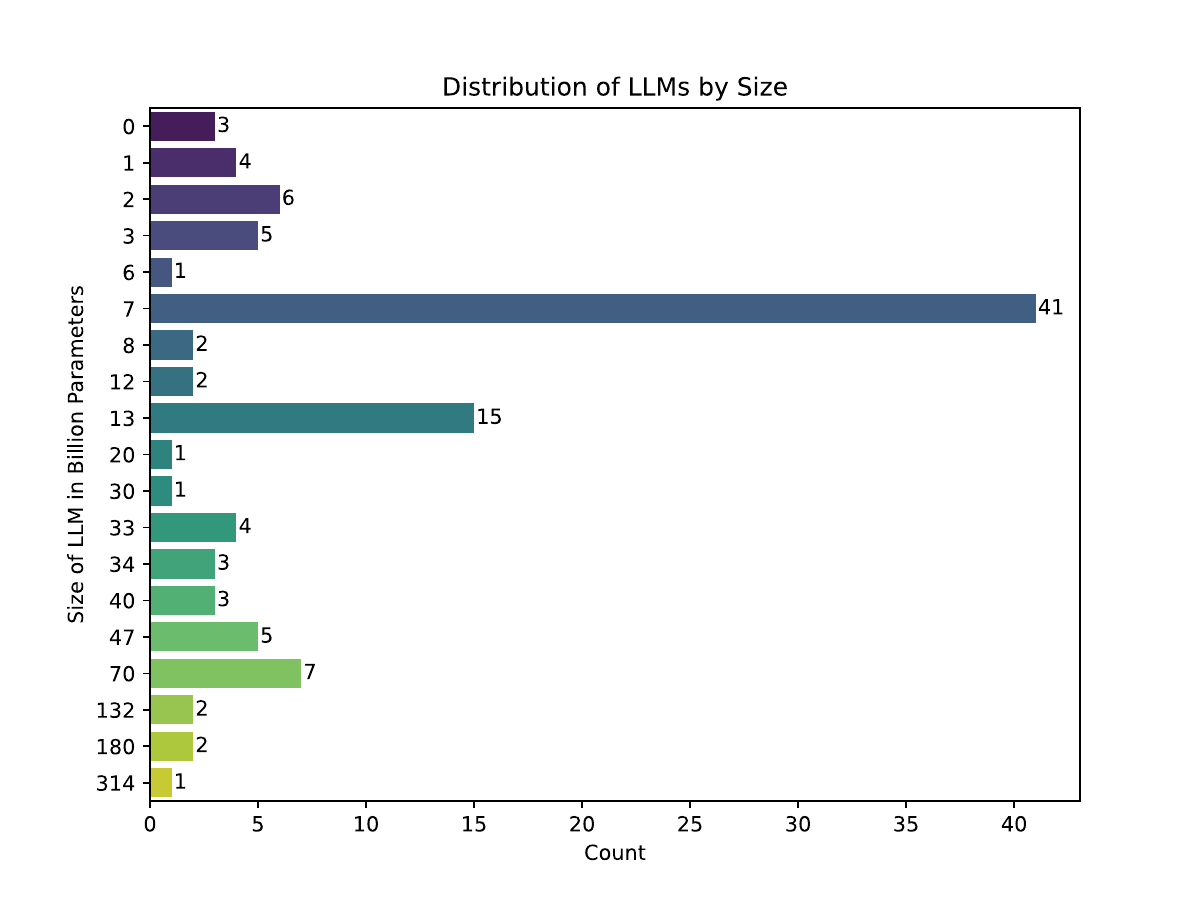}}
  \vspace{-20pt}
  \caption{Distribution of \ac{LLM} Size in Billion Parameters}
  \label{fig:size_distribution}
\end{figure}

Figure \ref{fig:size_distribution} shows the distribution of model size, indicating that; most of our listed \acp{LLM} have 7 billion parameters. The size of the rest of the models ranges from 13 billion to 314 billion parameters). The lower number of parameters can allow an \ac{LLM} to be deployed on edge devices, e.g., NVIDIA Jetson while the larger ones require more hardware resources. 
    
\begin{table}[htbp]
  \centering
  \caption{License Distribution of Open-Source Models in our List}
  \label{tab:license_distribution}
  \begin{tabularx}{\columnwidth}{|X|c|r|r|}
    \toprule
    \textbf{License Type} & \textbf{Count} & \textbf{Percentage (\%)} \\
    \midrule
    Apache 2.0 & 36 & 33.33 \\
    LLaMA-2 & 29 & 26.85 \\
    Gemma & 12 & 11.11 \\
    MIT & 7 & 6.48 \\
    CC-BY-NC 4.0 & 5 & 4.63 \\
    CC-BY-NC-ND 4.0 & 4 & 3.70 \\
    LLaMA-3 & 4 & 3.70 \\
    Non-commercial & 3 & 2.78 \\
    Microsoft Research License & 2 & 1.85 \\
    Databricks Open Model License & 2 & 1.85 \\
    Falcon-180B TII license & 2 & 1.85 \\
    Med42 (derivative of LLaMA-2) & 1 & 0.93 \\
    StabilityAI Non-Commercial Research Community License & 1 & 0.93 \\
    \midrule
    \textbf{Total} & 108 & --- \\
    \bottomrule
  \end{tabularx} 
\end{table}

Table \ref{tab:license_distribution} shows the distribution of license categories among our listed \ac{LLM} models. Regarding commercial usage of the listed \acp{LLM}, around 51\% of models have permissive licenses (Apache 2.0, MIT, Gemma) that allow for commercial usage without permission from model authors. Additionally, approximately 32\% of listed \acp{LLM} have limited commercial usage licenses (LLaMA-2, LLaMA-3, DataBricks Open Model License). Models with such licenses require permission from model authors if commercial usage exceeds 700M monthly active users. In Table \ref{tab:llm-table}, such models are denoted as ``Partial'' commercial usage. 

Our comparison table includes \acp{LLM} that have been specifically fine-tuned for the medical domain. Reducing hallucinations is particularly crucial in the medical field, as the generated responses may be used for diagnosis and treatment. Consequently, medical \acp{LLM} like BioMistral, Medicine-LLM, and Meditron have been fine-tuned by their developers using textual data from PubMed Central Open Access, internationally recognized medical guidelines, and a meticulously curated medical corpus. 

\section{Conclusion}
\label{sec: conclusion}
In this paper, we proposed a comprehensive list of \acp{LLM}. This list is aimed at supporting researchers and companies in selecting \ac{LLM} that is suitable for their use case, needs, and hardware requirements. This list is an ongoing effort and will be updated as new pre-trained or fine-tuned \acp{LLM} arrive. 

Fine-tuning capability of \acp{LLM} has lad to many derivations of them for specific use cases. Since listing every fine-tuned \ac{LLM} may not help researchers and companies and on the opposite; may confuse them more, this list does not cover all the fine-tuned versions of foundational \acp{LLM}. Another limitation is that the proposed list may not include the latest \acp{LLM} since the update frequency of the table may not align with the publication of new ones. 

In future work, we will include more domain-specific models to list the \ac{LLM} options for different applications. Furthermore, we will assess user feedback and highlight the advantages and disadvantages of the recommended deployments. Note that, in this study, the \acp{LLM} listed were not tested. The requirements provided by HuggingFace and the developers of \acp{LLM} will be verified as part of the future work. 

\section*{Acknowledgment}
We thank Nehal Darwish (University of Stuttgart, Institute of Industrial Manufacturing and Management (IFF)) for preparing the first draft of the comparison list. 

%%%%%%%%%%%%%%%%%%%%%%%%%%%%%%%%%%%%%%%%%%%%%%%%%%%%%%%%%%%%%%%%%%%%%%%%%%%%%%%%

\printbibliography[notcategory=selfref]

@article{chang.2024,
    author = {Chang, Yupeng and Wang, Xu and Wang, Jindong and Wu, Yuan and Yang, Linyi and Zhu, Kaijie and Chen, Hao and Yi, Xiaoyuan and Wang, Cunxiang and Wang, Yidong and Ye, Wei and Zhang, Yue and Chang, Yi and Yu, Philip S. and Yang, Qiang and Xie, Xing},
    title = {A Survey on Evaluation of Large Language Models},
    year = {2024},
    issue_date = {June 2024},
    publisher = {Association for Computing Machinery},
    address = {New York, NY, USA},
    volume = {15},
    number = {3},
    pages = {1-45},
    issn = {2157-6904},
    url = {https://doi.org/10.1145/3641289},
    doi = {10.1145/3641289},
    journal = {ACM Trans. Intell. Syst. Technol.},
    articleno = {39},
    numpages = {45}
}

@article{zhao2023survey,
      title={A Survey of Large Language Models}, 
      author={Wayne Xin Zhao and Kun Zhou and Junyi Li and Tianyi Tang and Xiaolei Wang and Yupeng Hou and Yingqian Min and Beichen Zhang and Junjie Zhang and Zican Dong and Yifan Du and Chen Yang and Yushuo Chen and Zhipeng Chen and Jinhao Jiang and Ruiyang Ren and Yifan Li and Xinyu Tang and Zikang Liu and Peiyu Liu and Jian-Yun Nie and Ji-Rong Wen},
      journal={arXiv preprint arXiv:2303.18223},
      year={2023},
      url={https://arxiv.org/abs/2303.18223}
}

@article{zhao2024_exp,
    author = {Zhao, Haiyan and Chen, Hanjie and Yang, Fan and Liu, Ninghao and Deng, Huiqi and Cai, Hengyi and Wang, Shuaiqiang and Yin, Dawei and Du, Mengnan},
    title = {Explainability for Large Language Models: A Survey},
    year = {2024},
    issue_date = {April 2024},
    publisher = {Association for Computing Machinery},
    address = {New York, NY, USA},
    volume = {15},
    number = {2},
    pages = {1-38},
    issn = {2157-6904},
    url = {https://doi.org/10.1145/3639372},
    doi = {10.1145/3639372},
    journal = {ACM Trans. Intell. Syst. Technol.},
    articleno = {20},
    numpages = {38}
}

@article{wang2023aligning,
      title={Aligning Large Language Models with Human: A Survey}, 
      author={Yufei Wang and Wanjun Zhong and Liangyou Li and Fei Mi and Xingshan Zeng and Wenyong Huang and Lifeng Shang and Xin Jiang and Qun Liu},
      journal={arXiv preprint arXiv:2307.12966},
      year={2023},
      url={https://arxiv.org/abs/2307.12966}
}

@article{zhang2024instruction,
  title={Instruction Tuning for Large Language Models: A Survey},
  author={Zhang, Shengyu and Dong, Linfeng and Li, Xiaoya and Zhang, Sen and Sun, Xiaofei and Wang, Shuhe and Li, Jiwei and Hu, Runyi and Zhang, Tianwei and Wu, Fei and others},
  journal={arXiv preprint arXiv:2308.10792},
  year={2024},
  url={https://arxiv.org/abs/2308.10792}
}

@article{gao2024retrievalaugmented,
      title={Retrieval-Augmented Generation for Large Language Models: A Survey}, 
      author={Yunfan Gao and Yun Xiong and Xinyu Gao and Kangxiang Jia and Jinliu Pan and Yuxi Bi and Yi Dai and Jiawei Sun and Meng Wang and Haofen Wang},
      journal={arXiv preprint arXiv:2312.10997},
      year={2024},
      url={https://arxiv.org/abs/2312.10997}
}

@article{huang2023reasoning,
      title={Towards Reasoning in Large Language Models: A Survey}, 
      author={Jie Huang and Kevin Chen-Chuan Chang},
      journal={arXiv preprint arXiv:2212.10403},
      year={2023},
      url={https://arxiv.org/abs/2212.10403}
}

@online{open-llm-leaderboard,
  author = {Edward Beeching and Clémentine Fourrier and Nathan Habib and Sheon Han and Nathan Lambert and Nazneen Rajani and Omar Sanseviero and Lewis Tunstall and Thomas Wolf},
  title = {Open LLM Leaderboard},
  year = {2023},
  publisher = {Hugging Face},
  note = {Accessed: 2024-05-28},
  url = {https://huggingface.co/open-llm-leaderboard}
}

@software{eval-harness,
  author       = {Gao, Leo and
                  Tow, Jonathan and
                  Biderman, Stella and
                  Black, Sid and
                  DiPofi, Anthony and
                  Foster, Charles and
                  Golding, Laurence and
                  Hsu, Jeffrey and
                  McDonell, Kyle and
                  Muennighoff, Niklas and
                  Phang, Jason and
                  Reynolds, Laria and
                  Tang, Eric and
                  Thite, Anish and
                  Wang, Ben and
                  Wang, Kevin and
                  Zou, Andy},
  title        = {A framework for few-shot language model evaluation},
  month        = sep,
  year         = 2021,
  publisher    = {Zenodo},
  version      = {v0.0.1},
  doi          = {10.5281/zenodo.5371628},
  url          = {https://doi.org/10.5281/zenodo.5371628}
}

@article{muennighoff2022mteb,
    doi={10.48550/arxiv.2210.07316},
    url={https://arxiv.org/abs/2210.07316},
    author={Muennighoff, Niklas and Tazi, Nouamane and Magne, Lo{\"\i}c and Reimers, Nils},
    title={MTEB: Massive Text Embedding Benchmark},
    publisher={arXiv},
    journal={arXiv preprint arXiv:2210.07316},  
    year={2022}
}

@article{chiang2024chatbot,
      title={Chatbot Arena: An Open Platform for Evaluating LLMs by Human Preference}, 
      author={Wei-Lin Chiang and Lianmin Zheng and Ying Sheng and Anastasios Nikolas Angelopoulos and Tianle Li and Dacheng Li and Hao Zhang and Banghua Zhu and Michael Jordan and Joseph E. Gonzalez and Ion Stoica},
      journal={arXiv preprint arXiv:2403.04132},
      year={2024},
      url={https://arxiv.org/abs/2403.04132}
}

@article{bradley-terry,
     ISSN = {00063444},
     URL = {http://www.jstor.org/stable/2334029},
     author = {Ralph Allan Bradley and Milton E. Terry},
     journal = {Biometrika},
     number = {3/4},
     pages = {324--345},
     publisher = {[Oxford University Press, Biometrika Trust]},
     title = {Rank Analysis of Incomplete Block Designs: I. The Method of Paired Comparisons},
     urldate = {2024-04-28},
     volume = {39},
     year = {1952}
}

@online{eugeneyan-github,
  title = {Open LLMs},
  author={Eugene Yan},
  url = {https://github.com/eugeneyan/open-llms},
  note = {Accessed: 2024-05-28}
}

@online{sapling-web,
  title = {The LLM Index},
  url = {https://sapling.ai/llm/index},
  note = {Accessed: 2024-05-28}
}

@misc{mixtral,
  author = {Mistral AI Team},
  title = {Cheaper, Better, Faster, Stronger: Continuing to push the frontier of AI and making it accessible to all.},
  year = {2024},
  month={May},
  publisher = {Mistral AI},
  url = {https://mistral.ai/news/mixtral-8x22b/}
}

@article{falcon,
  title={The Falcon Series of Language Models: Towards Open Frontier Models},
  author={Almazrouei, Ebtesam and Alobeidli, Hamza and Alshamsi, Abdulaziz and Cappelli, Alessandro and Cojocaru, Ruxandra and Alhammadi, Maitha and Daniele, Mazzotta and Heslow, Daniel and Launay, Julien and Malartic, Quentin and Noune, Badreddine and Pannier, Baptiste and Penedo, Guilherme},
  year={2023}
}

@online{GenAIModelOverview,
    author    = {Alper Yaman and Jannik Schwab and Christof Nitsche and Abhirup Sinha and Marco Huber},
    title     = {Gen-AI Model Overview Table},
    year      = {2024},
    url       = {https://technology-project-aimv-projects-generative-ai-54af1e2b8cbbab0a.pages.fraunhofer.de},
    urldate   = {2024},
    note = {Accessed: 2024-06-13}
}

@INPROCEEDINGS{selfref,
      title={{Comparison of Large Language Models for Deployment Requirements}},
      author={Alper Yaman and Jannik Schwab and Christof Nitsche and Abhirup Sinha and Marco Huber},
      booktitle={Proceedings of the First International Conference on Generative Pre-trained Transformer Models and Beyond (GPTMB 2024)},
      address = {Porto, Portugal},
      month=jun,
      year={2024},
      url = {https://www.thinkmind.org/index.php?view=article&articleid=gptmb_2024_2_30_30008},
      pages = {41--44}
    }
\end{document}